\documentclass[letterpaper, 10 pt, conference]{ieeeconf}
\usepackage[numbers,sort&compress]{natbib}
\usepackage{graphicx}
\usepackage{booktabs}
\usepackage{amsmath}
\usepackage{amssymb}
\usepackage{bbm}
\usepackage{caption}
\usepackage{xcolor}
\usepackage{soul}
\usepackage{tabularx}
\usepackage{colortbl}  %
\usepackage{array}
\usepackage{listings}
\usepackage{mathrsfs}
\usepackage{wrapfig}
\usepackage{vcell}
\usepackage{pifont} %
\usepackage{marvosym}   %
\usepackage[ruled,vlined]{algorithm2e}
\usepackage{acronym}
\usepackage[pagebackref,breaklinks,colorlinks,allcolors=ourcolor]{hyperref}
\usepackage[capitalize]{cleveref}

\crefname{algorithm}{Alg.}{Algs.}
\Crefname{algocf}{Algorithm}{Algorithms}
\crefname{section}{Sec.}{Secs.}
\Crefname{section}{Section}{Sections}
\crefname{table}{Tab.}{Tabs.}
\Crefname{table}{Table}{Tables}
\crefname{figure}{Fig.}{Fig.}
\Crefname{figure}{Figure}{Figure}

\definecolor{revision}{RGB}{0,0,255}
\definecolor{iccvblue}{rgb}{0.21,0.49,0.74}

\definecolor{Gray}{gray}{0.85}

\newcommand{\method}{\texttt{\textbf{Hi-ORS}}\xspace}
\newcommand{\methodfull}{Human-in-the-loop Online Rejection Sampling\xspace}
\definecolor{best}{rgb}{0.96, 0.57, 0.58}
\definecolor{second}{rgb}{0.98, 0.78, 0.57}
\definecolor{third}{rgb}{1.0, 1.0, 0.56}
\definecolor{shallowgreen}{RGB}{200, 230, 201} %
\definecolor{shalloworange}{RGB}{255, 224, 178} %
\definecolor{ourcolor}{RGB}{224, 33, 138}   % red pink
\definecolor{mypink}{RGB}{223,32,137}
\definecolor{myblue}{RGB}{32,137,223}

\acrodef{il}[IL]{Imitation Learning}
\acrodef{sft}[SFT]{Supervised Fine-tuning}
\acrodef{rl}[RL]{Reinforcement Learning}
\acrodef{mbrl}[MBRL]{Model-based Reinforcement Learning}
\acrodef{pomdp}[POMDP]{Partially Observable Markov Decision Process}
\acrodef{mdp}[MDP]{Markov Decision Process}
\acrodef{sde}[SDE]{Stochastic Differential Equation}
\acrodef{llm}[LLM]{Large Language Models}

\newcommand{\eg}{\textit{e.g.},\xspace}
\newcommand{\ie}{\textit{i.e.},\xspace}

\usepackage{amsmath,amsfonts,bm}

\def\eqref#1{equation~\ref{#1}}

\def\1{\bm{1}}

\DeclareMathAlphabet{\mathsfit}{\encodingdefault}{\sfdefault}{m}{sl}
\SetMathAlphabet{\mathsfit}{bold}{\encodingdefault}{\sfdefault}{bx}{n}

\def\gA{{\mathcal{A}}}

\def\gI{{\mathcal{I}}}

\def\gL{{\mathcal{L}}}

\def\gN{{\mathcal{N}}}

\def\gS{{\mathcal{S}}}

\def\sR{{\mathbb{R}}}

\newcommand{\E}{\mathbb{E}}

\title{\LARGE \bf
Human-in-the-loop Online Rejection Sampling for Robotic Manipulation
}

\author{Guanxing Lu$^{1,2,\star}$, Rui Zhao$^{2,\star \dagger}$, Haitao Lin$^{2}$, He Zhang$^{2}$, Yansong Tang$^{1,\ddagger}$\\
$^\star$Equal contribution \quad $^\dagger$Project lead \quad $^\ddagger$Corresponding author\\
$^1$ Tsinghua Shenzhen International Graduate School, $^2$ Tencent Robotics X 
\\ \href{https://hiors-project.github.io/}{\texttt{\textbf{hiors-project.github.io}}}
\vspace{-0.3cm}
}

\begin{document}

\maketitle

\begin{abstract}

Reinforcement learning (RL) is widely used to produce robust robotic manipulation policies, but fine-tuning vision–language–action (VLA) models with RL can be unstable due to inaccurate  value estimates and sparse supervision at intermediate steps. In contrast, imitation learning (IL) is easy to train but often underperforms due to its offline nature. In this paper, we propose \methodfull (\method), a simple yet effective post-training method that utilizes rejection sampling to achieve both training stability and high robustness. \method stabilizes value estimation by filtering out negatively rewarded samples during online fine-tuning, and adopts a reward-weighted supervised training objective to provide dense intermediate-step supervision. For systematic study, we develop an asynchronous inference–training framework that supports flexible online human-in-the-loop corrections, which serve as explicit guidance for learning error-recovery behaviors. Across three real-world tasks and two embodiments, \method fine-tunes a $\pi_0$ base policy to master contact-rich manipulation in just 1.5 hours of real-world training, outperforming RL and IL baselines by a substantial margin in both effectiveness and efficiency. Notably, the fine-tuned policy exhibits strong test-time scalability by reliably executing complex error-recovery behaviors to achieve better performance.

\end{abstract}

\section{Introduction}

% VLA is good, but requires post-training
Vision-language-action models (VLAs) \citep{pomerleau1988alvinn, rt12022arxiv, rt22023arxiv, team2024octo, kim2024openvla, black2024pi0visionlanguageactionflowmodel, liu2024rdt, trilbmteam2025carefulexaminationlargebehavior} have become a prevailing approach for robotic manipulation. These models are pre-trained on massive heterogeneous teleoperation datasets with substantial compute \citep{khazatsky2024droid, walke2023bridgedata, open_x_embodiment_rt_x_2023, bu2025agibot}, thus cannot be applied out of the box in real-world deployments without further post-training. Post-training of VLAs generally adopts an \ac{il} approach that maximizes the likelihood of expert actions in collected states. As a pure offline exploitation method, \ac{il} can suffer catastrophic failures due to compounding errors: a failure during real execution may drive the system into states not present in the offline dataset, causing the entire episode to fail \citep{dagger,kelly2019hg}.

% We need rl, and rl's problems
To this end, \ac{rl} incorporates online exploration during training, which has been shown to produce robust real-world manipulation policies \citep{luo2024serl,hilserl}. However, training VLAs with real-world \ac{rl} is notoriously unstable.
For instance, \ac{rl} methods often require environment-specific hyperparameter tuning and free exploration, which is impractical for high-capacity VLAs and costly under real-world data-collection constraints. 
% Introduction question of our method
This naturally raises the question: \emph{How can we achieve stable and flexible online post-training for VLAs in robotic manipulation tasks?} Meeting this demand poses significant challenges for existing post-training methods.

% The reasons
Generally, \ac{rl} objective can be viewed as maximizing the probability of high-value actions. We argue that instability of real-world VLA post-training with \ac{rl} stems from two sources: (1) inaccurate value estimation: \ac{rl} uses neural networks to approximate the action-value function, which is susceptible to overestimation, especially in high-dimensional action spaces (\eg, action chunking \citep{qchunking}). (2) inefficient supervision: VLAs often benefit from leveraging intermediate computations prior to final action prediction (\eg, iterative denoising in diffusion-based policies \citep{chi2023diffusion}), but \ac{rl} typically supervises only the final action, resulting in sparse learning signals. These issues are exacerbated by limited on-robot sample budgets and real-robot safety constraints that restrict aggressive exploration.

\begin{figure}[t]
    \centering
    % \fbox{
    %     \begin{minipage}[c][0.25\textheight][c]{0.45\textwidth}
    %       \centering{blank}
    %     \end{minipage}
    %   }
    \includegraphics[width=0.45\textwidth]{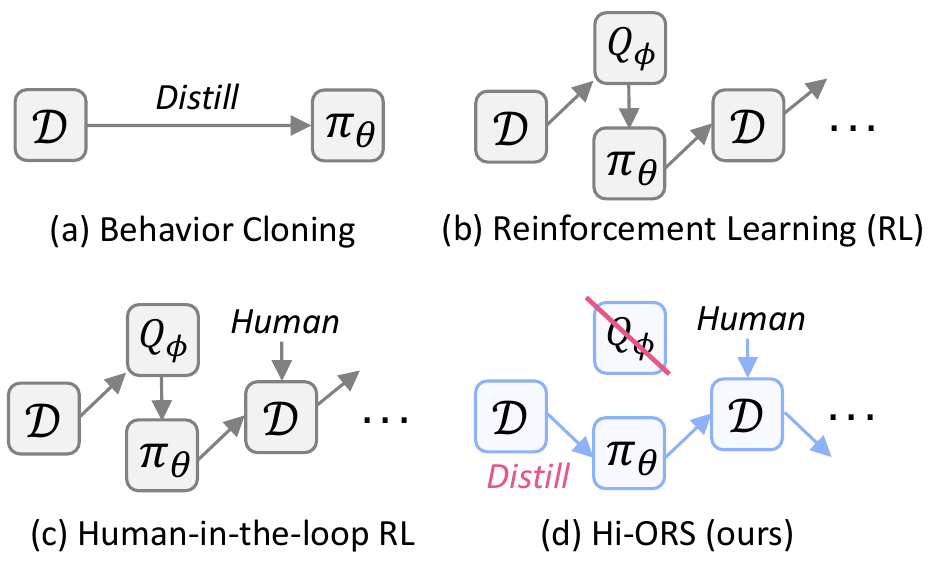}
    \caption{\small \textbf{\method} is a simple post-training method that stabilizes real-world RL. It replaces inaccurate value networks (\eg, in action chunking) with outcome-based rejection sampling, and implements a reward-weighted supervised training objective to distill dense intermediate-step supervision in VLAs (\eg, flow-matching–based). \method also incorporates online human-in-the-loop corrections as explicit guidance for learning error-recovery behaviors.
    }
    \label{fig:pipeline}
\end{figure}

% Human-in-the-loop
% Recent works \citep{hilserl} attempt to incorporate human intervention in real-world \ac{rl}.
To tackle these challenges, we propose \methodfull (\method), a simple yet effective post-training method for VLAs that achieves stable online training across diverse real-world tasks. At its core is a rejection sampling objective with strong theoretical guarantees \citep{neal2003slice}, which has been widely adopted in \ac{llm} literature \citep{anthropichh,rest}. Rather than learning a high-variance value function, \method performs outcome-based filtering: it discards negatively rewarded rollouts and retains successful episodes as judged by a golden reward model. This reduces reliance on approximate Q-functions and mitigates overestimation bias. To provide dense supervision over intermediate inference steps, we employ a simple and general supervised learning loss that trains both the final action predictions and the intermediate representations (\eg, denoising steps for diffusion policies or token-level predictions for autoregressive policies). 
\method also seamlessly incorporates flexible human interventions during data collection, including teleoperated corrections, targeted resets, and brief corrective segments injected mid-trajectory. These interventions provide explicit guidance for error recovery and diversify the accepted buffer with near-miss and recovery behaviors that are rare in offline datasets.
In a diverse set of three real tasks with two embodiments, \method fine-tunes a base model $\pi_0$ to master a contact-rich task in just 1.5 hours of real-world training, outperforming \ac{rl} and \ac{il} baselines by a sizable margin in both effectiveness and efficiency. Notably, we show that the fine-tuned policy has strong test-time scalability, which can repeatedly re-execute complex error-recovery behaviors to increase the test performance.

\begin{figure*}[t]
    \centering
    % \fbox{
    %     \begin{minipage}[c][0.3\textheight][c]{0.9\textwidth}
    %       \centering{blank}
    %     \end{minipage}
    %   }
    \includegraphics[width=0.99\textwidth]{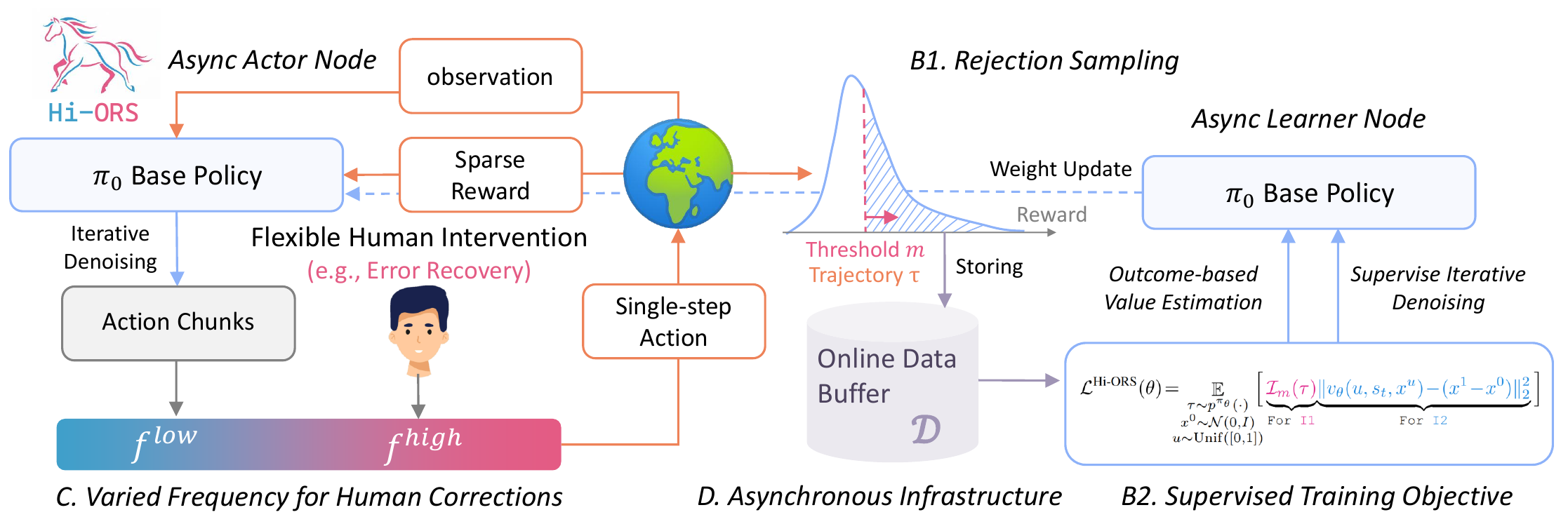}
    \caption{\small \textbf{The overall pipeline of \method}, which consists of a rejection sampling framework, a supervised training objective, a varied frequency strategy, and an asynchronous infrastructure. \method enables both training stability and high robustness in post-training VLAs for real-world robotic manipulation. Here we take a flow matching-based policy $\pi_0$ as an example.
    }
    \label{fig:pipeline}
\end{figure*}

In summary, our main contributions are threefold:
\begin{itemize}
    \item We first identify the crux of instability in \ac{rl} post-training for VLAs. Subsequently, we introduce \method, a simple and effective post-training method that stabilizes online learning via accurate outcome-based value estimation and a reward-weighted rejection sampling objective.
    \item We demonstrate that \method naturally incorporates human interventions to guide the policy in mastering error-recovery behaviors, yielding impressive test-time scalability.
    \item We validate \method on three challenging real-world tasks with two embodiments, improving upon \ac{il} and \ac{rl} baselines by large margins, while achieving high sample efficiency and minimal hyperparameter tuning.
\end{itemize}

% However, most existing deep RL algorithms entangle the choice of training objectives and algorithm design with the choice of the policy class. For example, soft actor-critic (SAC) \citep{sac}, the base learner for many offline and online fine-tuning algorithms \cite{rlpd}, employs reparameterization that is only stable for Gaussian policies:
% Swapping the policy for a diffusion policy causes instability \citep{wang2022diffusion}. 

\section{Related Work}
\label{sec:related_work}

% data
% \subsection{Real-world Post-training for VLAs}
\subsection{Imitation Learning for Robotic Manipulation}
Imitation learning \citep{zare2024survey, chebotar2023q} aims to recover expert strategies from given offline demonstrations. 
Among these methods, the most widely adopted variant is behavior cloning (BC) that maximizes the likelihood of expert actions, which has pushed the boundary of intelligent robots by decades \citep{pomerleau1988alvinn, rt12022arxiv, rt22023arxiv, team2024octo, kim2024openvla, black2024pi0visionlanguageactionflowmodel, liu2024rdt, trilbmteam2025carefulexaminationlargebehavior}. 
Growing large-scale robotic datasets \citep{khazatsky2024droid, walke2023bridgedata, open_x_embodiment_rt_x_2023, bu2025agibot} demonstrate that scaling demonstrations improves generalization of pre-trained policies on downstream tasks, as observed in other fields. Nevertheless, these pre-trained policies typically require online alignment (post-training) for sustained deployment in dynamic real-world environments.
% 略提human-in-the-loop
In this context, human-in-the-loop imitation learning \citep{dagger, kelly2019hg, wu2025robocopilothumanintheloopinteractiveimitation, mandlekar2020human, liu2022robot, hu2025racrobotlearninglonghorizon} collects interventions during on-policy rollouts to correct compounding errors and to expand coverage to failure states, enabling the agent to explore the unseen states and master new skills efficiently. 
For example, a recent work RaC \citep{hu2025racrobotlearninglonghorizon} leverages human-in-the-loop interventions for error recovery data collection, whereas it remains heavily dependent on human effort and lacks mechanisms for self-improvement. In contrast, our method maintains a supervised learning objective while enabling self-improvement.

% \subsection{Human-in-the-loop Robot Learning}
\subsection{Reinforcement Learning for Robotic Manipulation}
To support self-improvement, reinforcement learning is a post-training paradigm that optimizes actions via trial-and-error to maximize expected return.
However, applying \ac{rl} to real-world VLA training is non-trivial \citep{luo2024serl, hilserl}, which requires systematic infrastructure designs and is notoriously hard to train.
To mitigate the unstable training dynamics, recent works \citep{park2025flow,lu2025vla,guo2025improving,mark2024policy,qchunking} have explored temporal abstraction and hybrid objectives.
As an example, Q-chunking \citep{qchunking} introduces action chunking into temporal difference-based RL to improve temporal credit assignment, and iRe-VLA\citep{guo2025improving} alternates IL and RL to stabilize updates. Nevertheless, these methods are primarily validated in simulation.
Another recent work PA-RL \citep{mark2024policy} also leverages supervised training objectives to stabilize online training, but its action optimization strategy heavily relies on accurate value estimation and is in conflict with human-in-the-loop training.
Our method avoids unstable value-driven policy updates by an outcome-based rejection strategy, which shows impressive real-world performance in diverse task suites.

\subsection{Rejection Sampling}
The proposed method also connects to the broad literature on rejection sampling.
Rejection sampling \citep{neal2003slice} is a classical technique for drawing samples from a target distribution by filtering proposals.
In large language models, the term often refers to sampling multiple candidates and selecting the top-$k$ (or those that pass a verifier) for iterative self-improvement \citep{anthropichh, rest, gulcehre2023reinforced, zelikman2022star}.
For example, STaR~\citep{zelikman2022star} retrains on self-generated responses from the original pre-trained model that satisfy a verifier across iterations.
Our approach adapts this idea to real-world robot learning by performing reward-aware rejection of online rollouts. Combined with an asynchronous inference–training framework, this enables efficient incorporation of human-in-the-loop corrections and stabilizes post-training in contact-rich manipulation.

\section{\method}
\label{sec:approach}

In this section, we begin by formulating the problem of stabilizing reinforcement learning in VLA post-training and identifying the key challenges that motivate our approach (\Cref{subsec:preliminaries}). 
We then present \methodfull (\method), which leverages rejection sampling to achieve both training stability and high performance through filtered value estimation and reward-weighted supervision (\cref{subsec:method}). 
Next, we describe the incorporation of flexible online human-in-the-loop corrections, providing explicit guidance for learning complex error-recovery behaviors (\Cref{subsec:human}). 
Finally, we detail the implementation of our real-world robotic manipulation system that supports efficient online fine-tuning with human intervention (\Cref{subsec:system}).

\subsection{Preliminaries}\label{subsec:preliminaries}

We focus on robotic manipulation tasks in real-world domains, which can be defined by an \ac{mdp} expressed as the tuple $(\gS, \gA, p, \rho, r, \gamma)$. $\gS$ and $\gA=\sR^d$ refer to the state and $d$-dimensional continuous action space. For VLAs, states can be composed of multi-view images, natural language instructions, and optional proprioception. The action can be the next end-effector or joint pose trajectory, where a low-level planner acquires the motion. The state transition probability or environmental dynamics $p: \gS \times \gA \to \Delta(\gS)$ is unknown and potentially stochastic. $\rho$ is the initial state distribution. $r: \gS \times \gA \to \sR$ is the reward function, which is often sparse that only gives a positive value upon completion of the task. A scalar $\gamma$ denotes the discount factor.
We define the whole trajectory probability $p^{\pi_\theta}(\tau)$ as $\rho_0\left(s_0\right) \prod_{t=0}^{T-1} P\left(s_{t+1} \mid s_t, a_t\right) \pi\left(a_t \mid s_t\right)$. Then, the objective of \ac{rl} is to find the parameter $\theta$ of a policy $\pi_\theta$ to maximize the the average discounted return $R^{\pi_\theta} = \E_{\tau \sim p^{\pi_\theta}(\cdot)}[\sum_{t=0}^T \gamma^t r(s_t, a_t)]$ from online interaction experience, which is composed of multiple trajectories $\tau=(s_0, a_0, \ldots, s_T, a_T)$. 

To solve such \ac{mdp} problem involved in VLA post-training phase, a common practice is employing \ac{rl} algorithms. However, training VLAs with real-world \ac{rl} suffers from severe instability issues. Unfortunately, cutting-edge techniques in VLAs may intensify the instability. Thus, achieving stable and flexible online post-training for VLAs is demanding.
Without loss of generality, we analyze the classical policy gradient formulation to understand the root causes of this instability.
The classical policy gradient with respect to policy parameters $\theta$ is:
\begin{align}
\nabla_\theta \! \gL^{\text{PG}}(\theta) \!=\! - \!\mathbb{E}_{\tau \sim p^{\pi_\theta}(\cdot)} [Q_\phi(s,a) \nabla_\theta \log \pi_\theta(a_t|s_t)],
\label{eq:policy_gradient}
\end{align}
where $Q_\phi$ is another neural network serving as an approximation of the action-value function. 
This formulation reveals two primary sources of instability: 
\begin{enumerate}
    \item \textbf{\textcolor{mypink}{I1}}: inaccurate value estimation. $Q_\phi(s_t,a_t)$, which is particularly problematic when the action space is high-dimensional (\eg, action chunking case); 
    \item \textbf{\textcolor{myblue}{I2}}: inefficient supervision. $\log \pi_\theta(a_t|s_t)$ focuses on the final action while ignoring the intermediate compute prior to final action prediction, which is of great significance in current VLAs (\eg, denoising steps for diffusion policies or token-level generation for autoregressive policies).
\end{enumerate}
For instance, $\pi_0$ \citep{black2024pi0visionlanguageactionflowmodel} employs action chunking to predict multi-step action sequences and uses flow matching \citep{lipman2022flow} for continuous action generation.
Action chunking exponentially expands the action space with respect to the prediction horizon, making accurate value estimation significantly more difficult for $Q_\phi$.
Flow matching policies train a state- and time-dependent vector field $v_\theta(t, s, x): [0, 1] \times \gS \times \sR^d \to \sR^d$ that generates actions by solving an ODE from noise $x^0 \sim \gN(0, I_d)$ to target action $x^1 \equiv a$. The standard flow matching objective is:
\begin{align}
\gL^{\text{Flow}}(\theta)
\!=\! \mathop{\E}\limits_{\substack{\tau \sim p^{\pi_\theta}(\cdot)\\ x^0 \sim \gN(0, I_d)\\ u \sim \mathrm{Unif}([0,1])}}
\!\Big[
\big\| v_\theta(u, s_t, x^u) \!-\! (x^1\! -\! x^0) \big\|_2^2 \Big],
\label{eq:flow_matching}
\end{align}
where $x^u = (1-u)x^0 + ux^1$ represents the interpolation path.
Obviously, training flow matching with \ac{rl} requires iterative denoising with back-propagation through time (BPTT), which substantially increases variance and computational overhead during policy updates.
These challenges motivate our \method approach, which addresses value estimation inaccuracy through rejection sampling while providing dense supervision via a reward-weighted supervised objective to stabilize training.

% \subsection{High-Dimensional Instability of Policy Gradient}
\subsection{Rejection Sampling for Robotic Manipulation}\label{subsec:method}

Inspired by recent advances in LLM post-training \citep{anthropichh, rest, gulcehre2023reinforced, zelikman2022star}, we propose to utilize rejection sampling to overcome the challenges mentioned in the last subsection.
Unlike policy gradient methods that rely on learned value functions $Q_\phi$, rejection sampling provides a non-parametric approach to identify high-quality trajectories, remaining stable even in high-dimensional action spaces.
\method follows a two-phase structure analogous to Generalized Policy Iteration (GPI): the evaluation phase generates and filters trajectories from the current policy, while the improvement phase updates the policy using accepted high-reward samples. We can maintain off-policy data in the training mixture to prevent policy divergence from the base model.

\subsubsection{Evaluation Phase}\label{subsubsec:evaluation_phase}

The evaluation phase generates trajectories $\tau = (s_0, a_0, s_1, a_1, \ldots, s_T)$ from the current policy $\pi_\theta$ (or another exploration policy) and applies reward-based filtering to identify successful behaviors. Given a trajectory with cumulative reward $R(\tau) = \sum_{t=0}^{T} r_t$, we define an acceptance criterion using an indicator function:
\begin{equation}
\mathcal{I}_m(\tau) = \mathbbm{1}_{R(\tau) \geq m}
\label{eq:acceptance_indicator}
\end{equation}
where $m$ is a reward threshold that increases over training iterations. This filtering mechanism serves as our rejection sampling strategy, where trajectories below the threshold are rejected, and only high-performing trajectories are retained for policy updates.
The key insight is that by directly filtering based on task rewards rather than learned value estimates, we avoid the overestimation of $Q_\phi(s_t,a_t)$ in high-dimensional action spaces. Each accepted trajectory represents a genuine success, providing reliable supervision for policy improvement. 

\subsubsection{Improvement Phase}\label{subsubsec:improvement_phase}

After filtering, we update the policy using a reward-weighted supervised learning objective that mimics successful behaviors for its simplicity:
\begin{align}
\nabla_\theta \gL^{\text{Hi-ORS}}(\theta) \!=\! - \!\mathbb{E}_{\tau \sim p^{\pi_\theta}(\cdot)} [\gI_m(\tau)  \nabla_\theta \log \pi_\theta(a_t|s_t)],
\label{eq:improvement_phase}
\end{align}
% where $\mathcal{D}$ represents the online trajectory dataset.
We now illustrate how to use this training objective to supervise intermediate inference steps in modern VLAs.
For flow matching-based VLAs \citep{chi2023diffusion, black2024pi0visionlanguageactionflowmodel}, we convert rejection-sampled trajectories into dense supervision for the vector field at all intermediate integration times. Given the acceptance indicator $\gI_m(\tau)$ from \Cref{eq:acceptance_indicator}, we extract time-indexed pairs $(s_t, a_t)$ from generated correct trajectories. For each accepted pair, we optimize:
\begin{align}
\gL^{\text{Hi-ORS}}(\theta)
\!=\!\!\!\! \mathop{\E}\limits_{\substack{\tau \sim p^{\pi_\theta}(\cdot)\\ x^0 \sim \gN(0, I)\\ u \sim \mathrm{Unif}([0,1])}}
\!\!\!\!\Big[\underbrace{\color{mypink}\gI_m(\tau)}_{\color{black}\texttt{For \textcolor{mypink}{I1}}} \!
\underbrace{\color{myblue}\| v_\theta(u, s_t, x^u) \!-\! (x^1\! -\! x^0) \|_2^2}_{\color{black}\texttt{For \textcolor{myblue}{I2}}} \color{black} \Big],
\label{eq:flow_hiors}
\end{align}
In \Cref{eq:flow_hiors}, the first term is the indicator function that performs stable value estimation, and the second term is the flow matching loss that provides dense supervision across denoising times $u$, addressing both key sources of instability in \Cref{eq:policy_gradient}.

If we sample from improved policies, the average reward of the generated samples would increase.
To ensure continuous improvement, we can implement a progressive threshold schedule: $m_1 \leq m_2 \leq \cdots \leq m_N$ across $N$ training iterations. This filtering with the growing threshold results in data subsets of increasing quality but of decreasing size. Consecutive fine-tuning of policies $\{\pi_{\theta_k}\}_{k\ge1}$ on higher quality data subsets ensures monotonic policy improvement.  

In practice, evaluation and improvement phases run asynchronously, enabling efficient off-policy learning where separate policy copies handle exploration and training. This design accommodates the computational overhead of VLA inference while maintaining stable learning. The update-to-data (UTD) ratio can be adjusted based on available computational resources (\Cref{subsec:system}).

\subsection{Varied Frequency for Human Corrections}\label{subsec:human}

The sample complexity of policy learning scales exponentially with state-action dimensionality and task horizon \citep{hilserl}. For complex manipulation tasks with high-dimensional visual observations and continuous action spaces, purely autonomous exploration becomes prohibitively expensive in real-world settings.
To address this challenge, we incorporate strategic human intervention that serves two critical purposes. The first is efficient exploration guidance by directing the policy toward promising regions of the state space. The second is explicit error recovery demonstration by showing the robot how to recover from failure modes that are difficult to discover autonomously.
During autonomous rollouts, \method supports a human operator to intervene at any timestep using relative end-effector control or absolute joint control.
Multiple interventions can occur within a single trajectory, creating mixed autonomous-human episodes. Critically, we only retain intervention episodes that achieve positive rewards according to our filtering criterion $\mathcal{I}_m(\tau)$ from \Cref{subsubsec:evaluation_phase}. This ensures that suboptimal human corrections do not contaminate the training data.
The key insight is that human interventions provide counterfactual demonstrations, showing the policy what it should have done in states where it was about to fail. This creates rich supervision for learning error recovery behaviors that would be nearly impossible to discover through random exploration.

To maximize data efficiency while maintaining execution quality, \method employs adaptive interaction frequency based on the control authority:
\begin{align}
f_{t} = \begin{cases}
f^{\text{high}}, & t \in \text{human intervention period;} \\
f^{\text{low}}, & t \in \text{autonomous control period,}
\end{cases}
\end{align}
where $f^{\text{high}} > f^{\text{low}}$ represents logging frequencies. During human intervention, we log transitions at a higher frequency to capture fine-grained corrective behaviors. During autonomous execution, we use a lower frequency to ensure consistent policy execution and avoid jerky motions or backtracking behaviors.

\subsection{Asynchronous Infrastructure}\label{subsec:system}

Given $G$ GPUs, we reserve one GPU for online inference and use the remaining $G-1$ GPUs for learning. An actor node streams data to a learner node, and multiple learners update the model via agentlace, following \citep{hilserl}. The learner is orchestrated with ZeRO-2 to enable large-scale distributed training for high-capacity VLAs. This asynchronous actor-learner design improves training throughput by about 2× and allows learning to continue even when the robot arm is halted, which is a common case in long-term real-world runs.
% We use bfloat16 precision to fit the model into GPU memory. 
For training stability, we filter no-op actions when the norm of the relative transform falls below a threshold to avoid initial stucks and discard very short episodes to prevent incorrect action chunking. 
The total latency consists of three parts, including model inference latency ($\sim160$ms), communication latency ($\sim400$ms), and sequential execution time ($\sim900$ms). The training time of one iteration is $\sim1.5$s, so the natural UTD ratio is around $1$. Under action chunking, the typical times of inference are $\sim20$ steps, resulting in $\sim20$s per episode.

\section{Experiments}
\label{sec:experiments}

\begin{figure}[t]
    \centering
    \begin{minipage}[c]{0.21\textwidth}
        \centering
        \includegraphics[width=\linewidth]{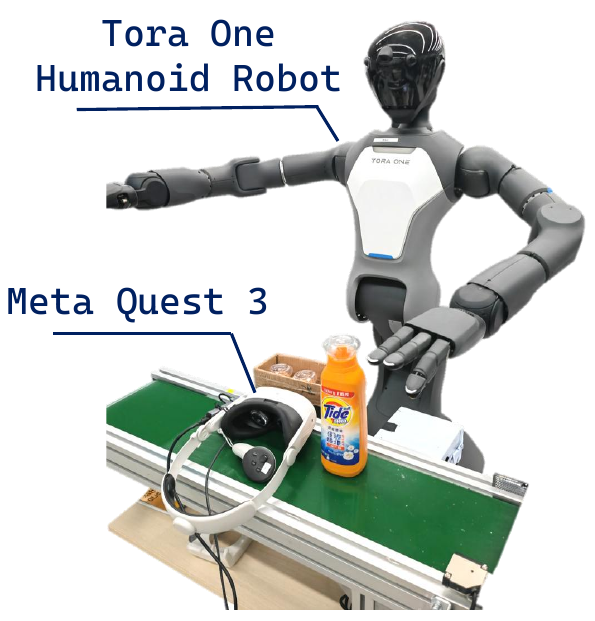}
    \end{minipage}\hfill
    \begin{minipage}[c]{0.27\textwidth}
        \centering
        \includegraphics[width=\linewidth]{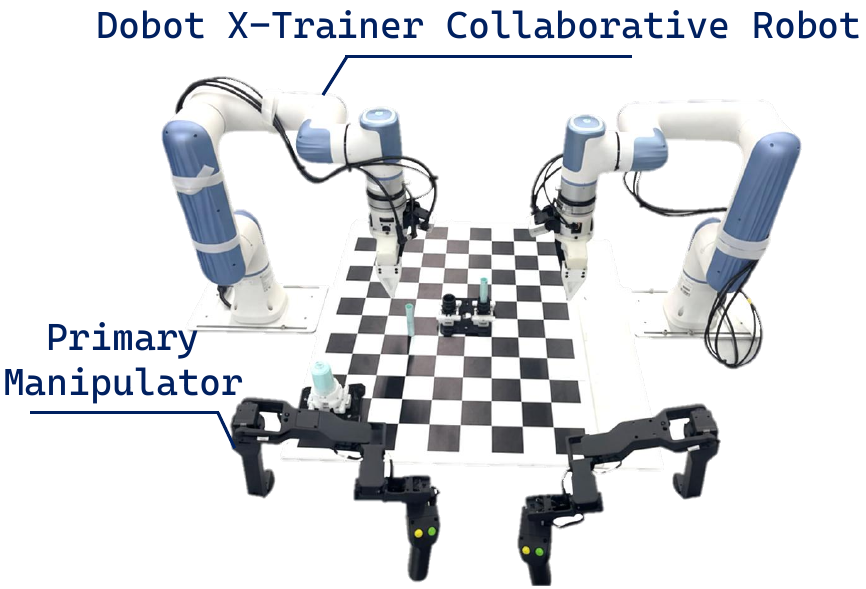}
    \end{minipage}
    \caption{\small \textbf{Real-world Settings}, we design three real-world tasks across two embodiments with different challenging levels to systematically evaluate the proposed method.}
    \label{fig:real_world_setups}
\end{figure}

In our experiments, we address the following questions:
\begin{enumerate}
    \item Q1: Does \method outperform prior methods in real-world robotic manipulation, in terms of effectiveness and efficiency?
    \item Q2: What are the learning dynamics of \method?
    \item Q3: How does each technique contribute to the overall performance of \method?
\end{enumerate}
In the following sections, we detail the model performance with respect to these questions.
We evaluate three tasks in two testing environments:
\begin{enumerate}
    \item \texttt{Raise-Hand}: a Paxini Tora One robot is instructed to raise its left arm to a target pose. The action space comprises the absolute end-effector pose and the gripper openness of the left arm. Human intervention is provided via a Meta Quest 3;
    \item \texttt{Pack-Detergent}: a Paxini Tora One robot is instructed to pick up laundry detergent from a conveyor belt and place it in a cardboard box;
    \item \texttt{Insert-Moisturizer}: a Dobot X-Trainer robot arm must pick up a thin moisturizer and insert it into the base. The action space comprises absolute joint angles and gripper openness. Intervention is provided by a primary arm via joint mapping. For all tasks, the observation space consists of images from the top and left wrist cameras, proprioception, and the task instructions.
\end{enumerate}
\Cref{fig:real_world_setups} shows the real-world setup of our experiments.
Our baselines include vanilla offline \ac{il} method behavior cloning, a widely-used real-world \ac{rl} method HIL-SERL \citep{hilserl} that incorporates value-based \ac{rl} with human-in-the-loop corrections, and a recent \ac{rl} method Q-Chunking \citep{qchunking} designed for action chunking. We use these compared methods to post-train a flow matching-based foundation VLA $\pi_0$ \citep{black2024pi0visionlanguageactionflowmodel} in all tasks. As behavior cloning, HIL-SERL, and Q-Chunking assume offline data, we collect initial human demonstrations for all counterparts. For simplicity, we manually annotate binary rewards rather than using a learned reward model. For evaluation, we randomly reset the environment and perform $10$ trials for each data point. 
$\pi_0$ is a widely used VLA using PaliGemma-3B \citep{beyer2024paligemma} as backbone and 300M parameters action expert for flow matching-based action chunk prediction.

\begin{figure*}[t]
    \centering
    % \fbox{
    %     \begin{minipage}[c][0.3\textheight][c]{1.0\textwidth}
    %       \centering{blank}
    %     \end{minipage}
    %   }
    \includegraphics[width=0.85\textwidth]{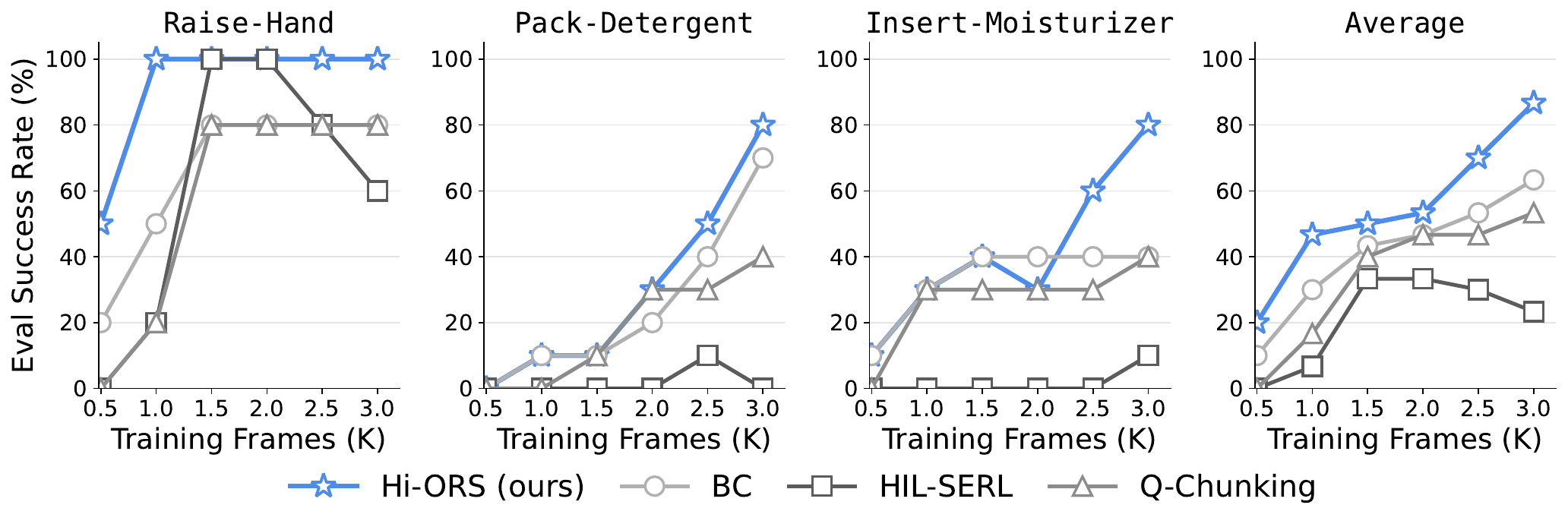}
    \caption{\small \textbf{Real-world Results}. We report the evaluation success rate curve of different methods in three real-world robotic manipulation tasks with different embodiments.
    }
    \label{fig:real_world_results}
\end{figure*}

\subsection{Real-world Experiments}\label{subsec:main_exp}

\subsubsection{Limits on Real-world RL for VLA}

In \Cref{fig:real_world_results}, we observe that HIL-SERL achieves strong performance and finds the optimal action (\ie, directly reaching the target pose) on a relatively simpler \texttt{Raise-Hand} task, despite showing instability with oscillatory regressions as updates continue.
However, harder tasks such as \texttt{Pack-Detergent} and \texttt{Insert-Moisturizer} pose challenges to the convergence of real-world \ac{rl} with VLAs.
While Q-Chunking stabilizes the performance by adding a distillation loss, it hinders further improvement.
We hypothesize that the inaccurate value estimation in high-dimensional chunked action space, and the BPTT issue in the gradient computation cause the instability of \ac{rl}. These effects jointly undermine real-world RL stability, motivating a value-free alternative that still provides dense supervision over intermediate inference steps.

\subsubsection{Compare \method with Previous Baselines}

\Cref{fig:real_world_results} shows that \method consistently outperforms prior baselines across all three real-world tasks, converging faster and attaining higher final success. 
% HIL-SERL
Relative to HIL-SERL, \method avoids value-function overestimation by replacing critic updates with a rejection-sampling evaluation phase, and then leverages accepted rollouts to provide dense, reward-weighted supervision of the flow field at all integration times. 
On \texttt{Raise-Hand}, \method matches the best performance of HIL-SERL but without late-stage regressions; on \texttt{Pack-Detergent} and \texttt{Insert-Moisturizer}, it reaches higher asymptotic success and requires fewer interactions to hit target success levels.
% Q-Chunking
Regarding Q-chunking, we incorporate human-in-the-loop corrections for fair comparisons. By incorporating the distillation loss to achieve intentional exploration in high-dimensional action space, it surpasses the HIL-SERL baseline but still underperforms \method. In this case, the advantage of \method stems from the effective outcome-based value estimation.
% IL
Compared to offline IL baselines, \method benefits from online data collection and the acceptance filter, mitigating compounding errors of offline methods. 
As a result, \method demonstrates improvements over behavior cloning with a sizable margin of $23.3$\% on average.
The performance drop of \method on \texttt{Insert-Moisturizer} with 2K frames results from the newly-added error recovery demonstrations, which exhibit different behavior patterns but facilitate subsequent learning.
Overall, the results validate that rejection sampling paired with reward-weighted flow supervision provides a stable, scalable post-training recipe for VLAs in real-world manipulation.

\begin{figure}[t]
    \centering
    % \fbox{
    %     \begin{minipage}[c][0.3\textheight][c]{0.4\textwidth}
    %       \centering{blank}
    %     \end{minipage}
    %   }
    \includegraphics[width=0.4\textwidth]{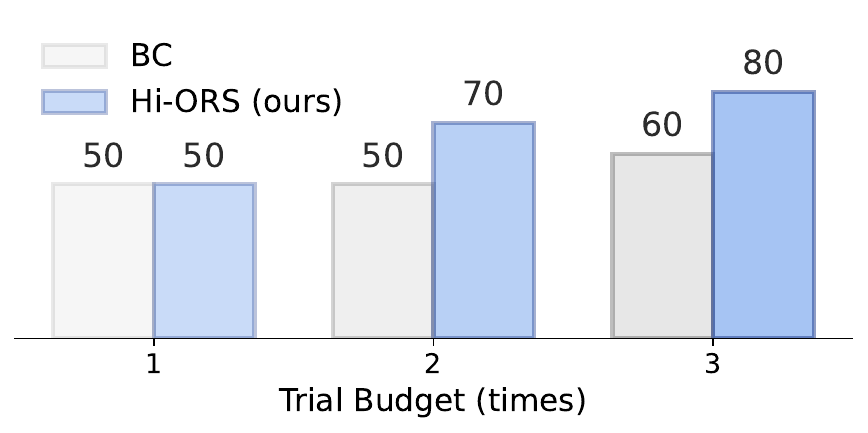}
    \caption{\small \textbf{Test-time Scaling} in \texttt{Insert-Moisturizer}. We show that larger trial budgets in evaluation result in higher testing performance, which indicates a potential signal of test-time scaling in robotic manipulation.
    }
    \label{fig:test_time_scaling}
\end{figure}

\subsection{Learning Dynamics}\label{subsec:learning_dynamics}

% \subsubsection{Episodic Length}
By performing reliable error recovery actions, \method indicates a potential path towards test-time scaling in robotic manipulation. To verify this, we evaluate the final checkpoint of \method and behavior cloning with different trial budgets.
In \Cref{fig:test_time_scaling}, \method exhibits clear test-time scaling.
This monotonic improvement indicates that the policy effectively uses additional retries to recover from intermediate errors rather than repeating failures. 
Besides, the marginal utility of increasing test-time compute is diminishing, as shown in the figure.
In contrast, behavior cloning policies show little scaling effect, suggesting limited capacity for purposeful recovery at test time.

\subsection{Spatial Generalization}\label{subsec:spatial_generalization}

\begin{figure}[t]
    \centering
    % \fbox{
    %     \begin{minipage}[c][0.3\textheight][c]{0.4\textwidth}
    %       \centering{blank}
    %     \end{minipage}
    %   }
    \includegraphics[width=0.49\textwidth]{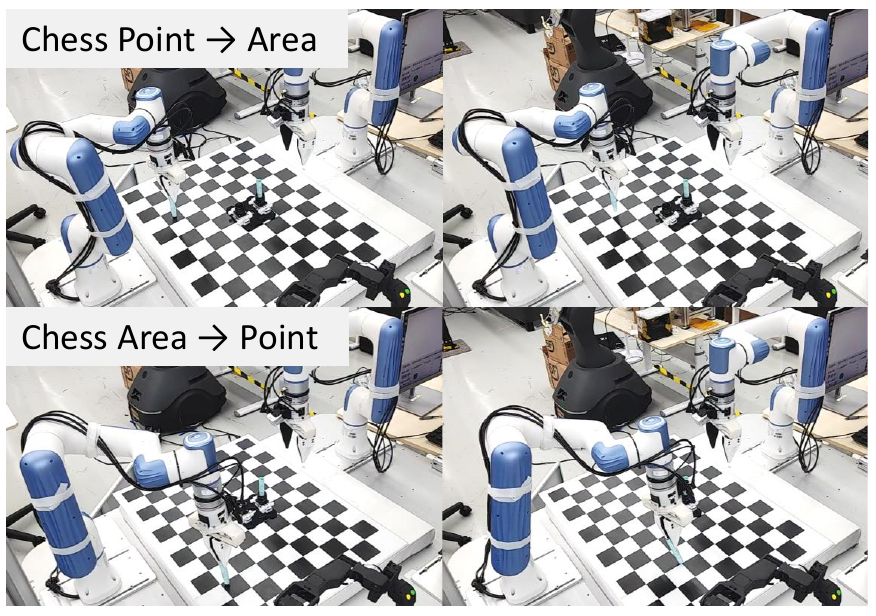}
    \caption{\small \textbf{Spatial Generalization}. We shows four extreme cases to validate the spatial generalizability of the proposed method.
    }
    \label{fig:spatial_generalization}
\end{figure}

In this subsection, we evaluate the spatial generalizability of \method with a curriculum data collection strategy. We first collect data where the object is initially located in chess points by human intervention, and evaluate \method on test cases where the object is located in the middle area of the chess grid. Then we conduct similar experiments with different training and testing cases. The results in \Cref{fig:spatial_generalization} show that \method exhibits great spatial generalizability even in extreme cases where the object is located far away from the robot's reset home. This generalizability contributes to the online data collection nature of \method, which enables quick fix of manipulation policies.

\subsection{Error Recovery}\label{subsec:error_recovery}

\begin{figure*}[t]
    \centering
    % \fbox{
    %     \begin{minipage}[c][0.2\textheight][c]{1.0\textwidth}
    %       \centering{blank}
    %     \end{minipage}
    %   }
    \includegraphics[width=1\textwidth]{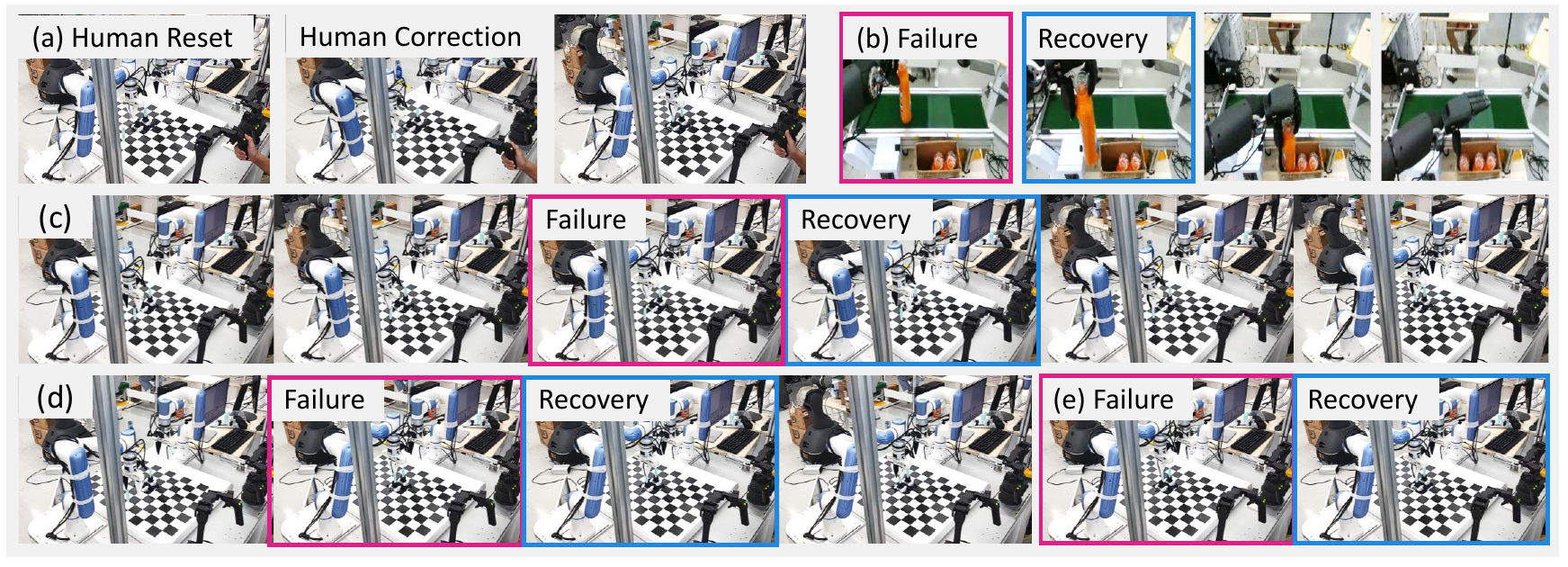}
    \caption{\small \textbf{Error Recovery Behaviors}. We show how human corrections enable fast error recovery mastering in sub-figure (a), and illustrate three impressive error recovery behaviors of \method in two robotic manipulation tasks, boosting its robustness in real-world deployments in sub-figures (b-e).
    }
    \label{fig:error_recovery}
\end{figure*}

% error behavior
\Cref{fig:error_recovery} illustrates several complex error-recovery behaviors performed by \method, along with a typical human correction strategy that enables rapid mastery of these skills. The behaviors include returning to re-grasp the object, lifting the gripper to reinsert the object, and performing a compensating insertion when the grasp pose is suboptimal. We also observe that an offline behavior cloning model fails quickly in similar cases, which helps explain the limited test-time scaling of the behavior cloning variant shown in \Cref{fig:test_time_scaling}.

\subsection{Ablation Studies}\label{subsec:ablation_studies}

\begin{table}[t]
\small
\setlength{\tabcolsep}{1pt}
\renewcommand{\arraystretch}{1.05}
\centering
\label{table:ablation_study} 
\begin{tabular}{l|cc}
\toprule
Hyperparameter & Training Time (h) & Success Rate (\%) \\
\midrule
\rowcolor{gray!10}
\textbf{\method} & \textbf{1.5} & \textbf{80} \\  
Cyclical Scheduler~\citep{smith2017cyclical} & 1.3 & 50 \\ 
Reward Classifier~\citep{hilserl}  & 1.5 & 60 \\ 
Remove Human Correction & - & 0 \\ 
Remove No-ops Action Filter & 2.2 & 20 \\ 
Remove Short Episode Filter & 1.5 & 60 \\ 
$5$-Step Execution & 1.0 & 10 \\ 
$25$-Step Execution & 1.5 & 40 \\ 
\bottomrule 
\end{tabular}
\caption{\small \textbf{Ablation Study.} We show the final average success rates on \texttt{Insert-Moisturizer}.
Removing any single technique from \method results in rapid collapse, emphasizing the essential role of each technique incorporated in \method.
}
\label{table:ablation}
\end{table}

We ablate major design choices in \method; results are summarized in \Cref{table:ablation}.

\paragraph{Choice of learning scheduler.}
We initially hypothesize that a cyclical scheduler~\citep{smith2017cyclical} may improve the training time as a higher learning rate fits new data faster while lower learning rate can help converge. However, our ablation experiments show that a cyclical scheduler has minor effect on both the training time and success rate. Based on Occam's Razor, we remove it in the final version of \method.

\paragraph{Choice of reward model.}
Replacing the human-annotated reward with a learned reward classifier~\citep{hilserl} yields a lower success rate with no training-time benefit. This is mainly because the reward model may predict false positive rewards in the process of human-in-the-loop error recovery, which confuses training.

\paragraph{Importance of human correction.}
% Human correction is important for learning error recovery behaviors, which are often hard to explore. 
By removing the human intervention, we observe an obvious performance drop in success rate as the model can not perform effective error recovery behavior to retry the evaluation task. 
Besides, removing human intervention also harms the training time, as the model may take an extremely long time to access positively rewarded samples without explicit guidance. 
The result validates the significance of human intervention in \method.

\paragraph{Choice of data filters.}
Both filters matter. Removing the no-ops action filter slows training and drops success to $20$\%, showing that pruning stuck transitions is essential for sample efficiency. This observation is aligned with \citep{kim2024openvla}. Removing the short-episode filter reduces success to $60$\%, indicating that trimming uninformative rollouts improves learning stability.

\paragraph{Varied execution frequency.}
The varied frequency strategy sets a high frequency during human inventions to obtain more data points, and sets a low frequency during model execution to avoid backtracking motions. Ablating the varied frequency by setting a fixed lower frequency (\eg, $5$-step) or a higher frequency (\eg, $25$-step) both result in performance degradation, confirming the effectiveness of the proposed varied frequency strategy.

Overall, \method achieves the best trade-off ($80$\% success rate within $1.5$h training). Each component contributes materially, as removing any single technique results in obvious performance drop.

% \paragraph{Environments and Tasks}
% To provide a comprehensive analysis of \method's capability, we evaluate our method on two real-world environments: 
% \section{Conclusion}
% \label{sec:conclusion}

% This work introduced \methodfull (\method), a simple online fine-tuning approach that stabilizes value estimation by filtering negatively rewarded samples, combining the training stability of IL with the robustness of RL. Across three real-world tasks and two embodiments, \method adapts a base policy $\pi_0$ to master contact-rich manipulation in about two hours of on-robot data, yielding sizable gains in both success rate and sample efficiency over strong RL and IL baselines. The resulting policies also exhibit test-time scalability, repeatedly executing complex error-recovery behaviors. These results position \method as a practical and interpretable baseline for real-world policy adaptation.
% % limitations
% Future work will extend to multi-task training, and xx. The threshold based filtering function may result into learning suboptimal behaviors that favors outcomes with high variance in the environments with stochastic dynamics.

% (Brandfonbrener et al., 2022).

\section{Conclusion}
\label{sec:conclusion}

We presented \methodfull (\method), a simple post-training method for VLAs that combines the robustness of RL with the stability of IL by using rejection sampling and reward-weighted supervision. \method filters out negatively rewarded samples to stabilize value estimation and trains a flow-matching policy with dense intermediate-step supervision. We further introduced an asynchronous inference–training framework with flexible online human-in-the-loop corrections that provide explicit guidance for learning error-recovery behaviors. Across three real-world tasks and two embodiments, \method adapts a base policy $\pi_0$ to contact-rich manipulation in about two hours of real-robot training, outperforming strong RL and IL baselines in both effectiveness and efficiency. The fine-tuned policies exhibit test-time scalability by reliably executing complex error-recovery behaviors. We advocate \method as a simple and robust baseline for fine-tuning VLAs in real-world robotic manipulation tasks.

Limitations and future work include extending \method to multi-task and longer-horizon settings, and improving the acceptance threshold scheduler in stochastic environments to avoid bias toward high-variance outcomes.

% \section*{Acknowledgement}

{
    \small
    \bibliographystyle{IEEEtran}
    \bibliography{reference}
}

\end{document}